\documentclass[pdflatex,sn-nature]{sn-jnl}

\usepackage{graphicx}%
\usepackage{multirow}%
\usepackage{amsmath,amssymb,amsfonts}%
\usepackage{amsthm}%
\usepackage{mathrsfs}%
\usepackage[title]{appendix}%
\usepackage{xcolor}%
\usepackage{textcomp}%
\usepackage{manyfoot}%
\usepackage{booktabs}%
\usepackage{algorithm}%
\usepackage{algorithmicx}%
\usepackage{algpseudocode}%
\usepackage{listings}%

\theoremstyle{thmstyleone}%
\theoremstyle{thmstyletwo}%

\theoremstyle{thmstylethree}%

\begin{document}

\title[Article Title]{Cross-level Requirement Traceability: A Novel Approach Integrating Bag-of-Words and Word Embedding for Enhanced Similarity Functionality}


\author*[1]{\fnm{Baher} \sur{Mohammad}}\email{baher.mohamad@hiast.edu.sy}

\author[1]{\fnm{Riad} \sur{Sonbol}}\email{riad.sonbol@hiast.edu.sy}

\author[1]{\fnm{Ghaida} \sur{Rebdawi}}\email{ghaida.rebdawi@hiast.edu.sy}

\affil*[1]{\orgdiv{Department of Informatics}, \orgname{Higher Institute for Applied Sciences and Technology (HIAST)}, \orgaddress{ \city{Damascus},  \country{Syria}}}


\abstract{Requirement traceability is the process of identifying the inter-dependencies between requirements. It poses a significant challenge when conducted manually, especially when dealing with requirements at various levels of abstraction. In this work, we propose a novel approach to automate the task of linking high-level business requirements with more technical system requirements.

The proposed approach begins by representing each requirement using a Bag-of-Words (BOW) model combined with the Term Frequency-Inverse Document Frequency (TF-IDF) scoring function. Then, we suggested an enhanced cosine similarity that uses recent advances in word embedding representation to correct traditional cosine similarity function limitations.

To evaluate the effectiveness of our approach, we conducted experiments on three well-known datasets: COEST, WARC(NFR), and WARC(FRS). The results demonstrate that our approach significantly improves efficiency compared to existing methods. We achieved better results with an increase of approximately 18.4\% in one of the datasets, as measured by the F2 score..}




\maketitle

\section{Introduction}\label{sec1}

A software requirement is a description of what the software will do and how it will be expected to perform \cite{bib1}. Requirements vary in degree of detail, starting from business requirements, which are the highest levels of abstraction that focus on the general purpose of the target system, all the way up to software requirements, which precisely describe the characteristics and features required from the system \cite{bib19}. Depending on these types, the task of requirements traceability varies between linking requirements from one level or linking requirements from different levels (such as linking business requirements to system requirements), which is called cross-level \cite{bib19}.

In general, requirements traceability is defined as “the ability to describe and follow the life of a requirement in both forwards and backward direction through periods of ongoing refinement and iteration” \cite{bib20}. This task plays a crucial role in the success of software development so that the development team can better understand the needs of customers and system users, and thus be able to design and develop a system that effectively meets those requirements \cite{bib16}.
Requirements traceability mainly aims to reveal the links and dependencies between software requirements, more specifically cross-level requirements traceability aims to ensure that each requirement at the high abstraction level is refined into a requirement at a lower level \cite{bib20}. Each low-level requirement (LLR) should be traced up to a specific high-level requirement (HLR); otherwise, subsequent design and implementation cannot satisfy system objectives or may exceed the system scope (over-standard) \cite{bib17}.

To address these issues, many studies tried to automate this task in the past two decades. Recently, many works \cite{bib2, bib9, bib10, bib12} adopted approaches to tackle this problem by representing requirements as vectors, mostly using Information Retrieval (IR) based methods or embeddings-based ones, and calculating similarity scores between each pair, then applying a threshold to determine similar requirements, However, these approaches did not manage to increase the efficiency, mainly because of the nature of the problem itself, as in cross-level requirements linking, the terms and vocabularies used in the high-level requirements and low-level ones are different in the degree of abstraction and details, which make this task more challenging than any other text similarity tasks, and demands special treatment \cite{bib20}. \newline

In the next section, we will provide a quick background about different approaches used to automate the task of connecting requirements, including both Information Retrieval (IR) and embedding based methods. We will introduce our approach in section 3, by discussing the intuition and motivation behind our work, as well as the detailed steps. Section 4 is dedicated to evaluating our approach on public datasets. Finally we will discuss the results in section 5, and conclude our paper in section 6.

\section{Related Work}\label{sec1}

To identify the most significant papers in this field, we based on a systematic mapping review conducted by one of the authors of this paper \cite{bib15}, and extended it to cover papers that are published after the covered period in that review.

We recognized two primary directions in related papers: IR-based approaches \cite{bib2,bib8} and embeddings-based approaches \cite{bib9,bib14}. 

Various methods have been employed among the IR-based approaches. Wang Et al. \cite{bib3} and Rempel et al. \cite{bib5} utilized the similarities between target artifacts to enhance their results. Wang et al. \cite{bib4} employed a hybrid approach combining BTM and TF-IDF techniques. Other studies employed IR to extract features and train machine learning models \cite{bib6,bib7}. Additionally, some approaches incorporated additional semantic information and utilized techniques such as LSI \cite{bib2} and ESA \cite{bib8}. 

On the other hand, many approaches suggested the use of embeddings-based methods, the most commonly used model was BERT. It was employed either in conjunction with a neural network classifier \cite{bib12,bib13} or by leveraging similarities between requirements \cite{bib9}. Guo Et al. \cite{bib14} trained an LSTM network to identify related requirements, while Waad et al. \cite{bib11} extracted semantic frames from requirements and employed the average of the pairwise similarities between these frames to identify possible links. Other researchers took a different direction by utilizing requirement-specific embeddings, by identifying the most important aspects of each requirement and the words associated with each aspect \cite{bib10}.

While these two approaches have been widely used, each one of them has its inherent limitations. The IR-based approach, specifically Bag-of-Words (BOW) combined with TF-IDF, when used with traditional similarity functions suffers from a lack of semantic information. This method represents each requirement within the unique word space of the corpus, assuming no relationships between words. Consequently, in this approach, each word is treated as an independent dimension perpendicular to other dimensions, which limits the ability to capture semantic nuances accurately. In contrast, embeddings-based approaches excel in representing words as vectors that effectively capture semantic information. However, these approaches face challenges when it comes to obtaining accurate document representations from individual word embeddings. Current pooling methods, such as average pooling, lack the necessary sophistication to capture the meaning of a document based on its constituent words.

\section{Proposed Method}\label{sec2}

we will divide our discussion into two subsections: (1) the main idea, where we will delve into the motivation behind adopting this approach, and (2) the detailed steps, where we will provide a comprehensive explanation of the solution's specifics.

\subsection{The main idea}\label{subsec2}

\begin{figure}[h]
    \centering
    \includegraphics[width=1\linewidth]{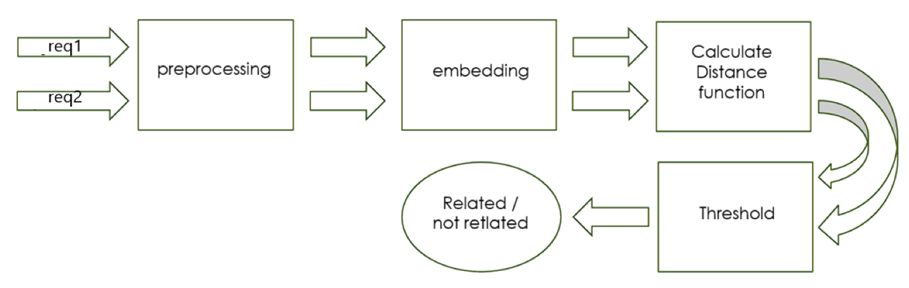}
    \caption{\centering the pipeline of the proposed approach}
    \label{fig:pipline}
\end{figure}

Our problem can be seen as a binary classification problem where the input is two requirements written in English (a pair of high-level and low-level requirements), and the output is whether these two requirements are related or not.  The pipeline of the system is composed of four main components as shown in figure \ref{fig:pipline}.

Previous works that follow this pipeline have primarily concentrated their efforts on the representation phase. They have explored two main directions, namely, IR-based approaches \cite{bib2,bib8,bib3,bib5,bib4,bib6,bib7} and embeddings-based approaches \cite{bib9,bib10,bib12,bib14,bib13,bib11}, as mentioned in the previous section. Additionally, they have utilized traditional similarity functions like cosine similarity or Manhattan distance.

In our work, we propose a hybrid solution that combines elements from both approaches. So instead of changing the representation method, we will introduce a new similarity function. Our choice to move away from traditional similarity functions, such as cosine similarity is driven by our objective to develop a novel function capable of capturing the interrelationships among dimensions (i.e., words in this representation method).\newline

For example, let's consider three documents: "use authentication," "add login," and "play football." If we employ the BOW+TF-IDF method to represent these documents, we will end up with something in table \ref{tab:TF-IDF}

\begin{table}[h]
    \centering
    \begin{tabular}{ccccccc}
         &  Use&  Authentication&  Add&  Login&  Play& Football
\\
         Doc1&   log(6)&   log(6)&  0&  0&  0& 0
\\
         Doc2&  0&  0&   log(6)&   log(6)&  0& 0
\\
         Doc3&  0&  0&  0&  0&   log(6)&  
log(6)\\
    \end{tabular}
    \caption{Representing different documents using BOW + TF-IDF}
    \label{tab:TF-IDF}
\end{table}

The limitations of traditional similarity functions, such as cosine similarity or Manhattan distance, become apparent when evaluating the similarity between document pairs that do not share matching terms. Despite the similarity between the first two documents and their dissimilarity from the third document, these traditional functions assign a similarity score of zero for all document pairs. This issue arises because these functions treat words as independent entities, overlooking the inherent relationships between them. 

To address this problem, we propose incorporating word similarities when calculating document similarity. Specifically, our proposed function takes into account the similarity matrix (similarities between every pair of words) alongside the two vectors being compared. Our proposed function must fulfill certain properties. \textbf{Firstly}, it should match the cosine similarity function when the similarity matrix is an identity matrix, indicating independence between dimensions. \textbf{Secondly}, it should be commutative, ensuring that the order of the vectors does not impact the similarity score. \textbf{Additionally}, the function should have a bounded range, allowing flexibility in adjusting the similarity threshold. \textbf{Finally}, the function should effectively handle scenarios similar to those illustrated in Table 1, where the similarity between the first and second documents is notably higher than other document similarities.

\subsection{The detailed steps}\label{subsec-detailed}
In the following, we discuss the details of each component of the proposed solution.
\begin{enumerate}
  \item Preprocessing: for this step, both high-level and low-level requirements are tokenized. Then we eliminate stopwords using a dictionary consisting of common stopwords in English and lemmatize words to convert them to their base forms.
  \item Representation: as mentioned above in the previous section, we will be using BOW+TF-IDF representation, and improve its results in the similarity function.
  \item Distance Function Formula: To formulate the new function, it is necessary to analyze the shortcomings of the cosine similarity in our specific case.
\end{enumerate}
 Upon examining the formula of the cosine function shown in equation \ref{eq:cosine_similarity}
\begin{equation}
\label{eq:cosine_similarity}
\text{cosine similarity}(\mathbf{A}, \mathbf{B}) = \frac{\sum_{i=1}^{n} A_i \cdot B_i}{\sqrt{\sum_{i=1}^{n} A_i^2} \cdot \sqrt{\sum_{i=1}^{n} B_i^2}}
\end{equation}
\newline

we observe that the dimensions are treated independently. The similarity between two vectors is calculated solely based on the similarity between the dimension $i$ from $A$ with the dimension $i$ only from $B$ , which corresponds to exact word matching between two documents, but it is crucial to consider the presence of related words and synonyms for a given word, especially in the case of cross-level requirements traceability, where the terms used in each level can be very different. Therefore, for each word in a document, we should account for the exact word match with the second document, as well as the impact of related and similar words on the overall vector similarity. This effect should be proportional to the similarity between the two words. Hence, the proposed function can be defined as the cosine similarity between $A$ and $sim$, where $sim$ is the matrix of pair wise similarities between the words of the corpus, so in this way we can incorporate the existence of similar words and synonyms. To maintain commutativity, the new function is formulated as equation \ref{eq:sim-func}
\begin{equation}
\label{eq:sim-func}
\text{similarity}(\mathbf{A}, \mathbf{B}, \text{sim}) = \frac{\cos(\mathbf{A}, \text{sim} \cdot \mathbf{B}) + \cos(\mathbf{B}, \text{sim} \cdot \mathbf{A})}{2}
\end{equation}
\newline
This function allows for the incorporation of word similarities into the calculation of document similarity. However, a potential issue may arise, as the exact word matching between two documents may be overlooked due to the cumulative impact of related words. In such cases, the effect of related words should augment exact word matching, rather than replacing it. 

To address this concern, we introduce two hyper-parameters: $similarity\_threshold$ and $synonym\_threshold$. The $similarity_threshold$ is a value below which any similarity between two words is considered negligible and set to 0. This prevents the accumulation of small similarities, particularly when document length increases. On the other hand, $the synonym\_threshold$ sets an upper limit on the combined impact of all related words for a given word. If the sum exceeds this threshold, the values are divided by a number that ensures this property. The division is performed as follows:

\begin{equation}
\label{eq:sim_update}
\text{sim}_{i,j} =
\begin{cases}
\frac{\text{synonym\_threshold} \times \text{sim}_{i,j}}{\sum_{j \neq i} \text{sim}_{i,j}}, & \text{if } \sum_{j \neq i} \text{sim}_{i,j} > \text{synonym\_threshold}\\
\text{sim}_{i,j}, & \text{otherwise}
\end{cases}
\end{equation}

\section{Experiments and Results}
\subsection{Dataset}
Our experiments were conducted on three datasets, MODIS, WARC(NFR) and WARC(FRS). We chose these data as all of them have two sets of requirements, high-level and low-level requirements written in free English without following any specific template. Table \ref{tab:datasets} shows the description and some statistics about the three datasets.

\begin{table}[h]
    \centering
    \begin{tabular}{|p{1.5cm}|p{3cm}|p{2cm}|p{2cm}|p{1cm}|p{1.2cm}|}
    
    \hline
    Dataset & Description & High-level requirements & Low-level requirements & Number of links & Source \\
    \hline
    MODIS   & A set of data from NASA Photography Moderate precision documentation Spectrometer & \centering 19 \newline requirements & \centering 49 \newline requirements & \centering 41 links & Promise Website \\
    \hline
    WARC(NFR) & A dataset for a web archive tool & \centering 21 \newline non-functional requirements & \centering 89 \newline software requirements specification & 58 links & Coest \\
    \hline
    WARC(FRS) & A dataset for a web archive tool & \centering 42 \newline functional requirements & \centering 89 \newline software requirements specification & 78 links & Coest \\
    \hline
    \end{tabular}
    \caption{\centering statistics about the used datasets }
    \label{tab:datasets}
\end{table}

\subsection{Evaluation}
The most important criteria to measure the success of any binary classification problem are precision and recall, as well as F-measure which is the weighted harmonic mean between them. In addition to these general-purpose standards, Hayes et al. \cite{bib18} introduced a practical standard specifically to the problem of candidate link generation in requirements based on their experience in the field, this standard reflects the fact that generated candidate links with high recall and low precision is better and more useful than those with high precision and low recall, this is because experts are better at checking whether a generated link is true than they are at generating new links. Table \ref{tab:levels} demonstrates the standard which is divided into 4 regions which are unacceptable, acceptable, good, and excellent.

\begin{table} [h]
    \centering
    \resizebox{\textwidth}{!}{%
    \begin{tabular}{|p{3cm}|p{3cm}|p{3cm}|}
    \hline
    Level & Recall & Precision \\
    \hline
    Excellent & Above 80\% & above 50\% \\
    \hline
    Good & 70\% - 80\% & 30\% - 50\% \\
    \hline
    Acceptable & 60\% - 70\% & 20\% - 30\% \\
    \hline
    Unacceptable & \multicolumn{2}{|c|}{else}\\
    \hline
    \end{tabular}%
}
    \caption{\centering different regions of acceptance}
    \label{tab:levels}
\end{table}

\subsection{Results}
We compared our proposed method with the methods found in similar works. All of these methods we experiment with are under the same schema as figure \ref{fig:pipline}, in other words, all of them represent the requirements, and decide whether they are related by looking at the similarity between them (after applying a certain threshold).
In Tables \ref{tab:MODIS} \ref{tab:WARC(NFR)} \ref{tab:WARC(FRS)} we compare the results of the proposed methods, with the ones used in similar works on the three datasets MOIDS, WARC(NFR), and WARC(FRS) respectively

\begin{table}[h]
\centering
\resizebox{\textwidth}{!}{%
\begin{tabular}{| l | l | l | l | l |}
\hline
\textbf{Method} & \textbf{Recall} & \textbf{Precision} & \textbf{$F_{1}$} & \textbf{$F_{2}$} \\
\hline
VSM \cite{bib3} & 36.6\% & 32.2\% & 33.7\% & 35.4\% \\
\hline
LSI \cite{bib2} & 70.7\% & 9.4\% & 16.6\% & 30.7\% \\
\hline
Fine-tuned BERT \cite{bib9} & 85.4\% & 10.4\% & 18.5\% & 34.9\% \\
\hline
Req2Vec \cite{bib10} & 41.5\% & 17.3\% & 24.4\% & 32.4\% \\
\hline
Ours & 73.2\% & 21.6\% & 33.3\% & 49.5\% \\
\hline
\end{tabular}%
}
\caption{\centering Results on MODIS}
\label{tab:MODIS}
\end{table}

\begin{table}[h]
\centering
\resizebox{\textwidth}{!}{%
\begin{tabular}{| l | l | l | l | l |}
\hline
\textbf{Method} & \textbf{Recall} & \textbf{Precision} & \textbf{Recall} & \\
\hline
VSM \cite{bib3} & 65.5\% & 27.9\% & 39.2\% & 51.6\% \\
\hline
LSI \cite{bib2} & 67.2\% & 25.3\% & 36.8\% & 50.5\% \\
\hline
Fine-tuned BERT \cite{bib9} & 36.2\% & 25.3\% & 29.8\% & 33.3\% \\
\hline
Req2Vec \cite{bib10} & 56.9\% & 37\% & 44.9\% & 51.4\% \\
\hline
Ours & 75.9\% & 53.7\% & 62.9\% & 70\% \\
\hline
\end{tabular}%
}
\caption{\centering Results on WARC(NFR)}
\label{tab:WARC(NFR)}
\end{table}

\begin{table}[h]
\centering
\resizebox{\textwidth}{!}{%
\begin{tabular}{| l | l | l | l | l |}
\hline
\textbf{Method} & \textbf{Recall} & \textbf{Precision} & \textbf{Recall} & \\
\hline
VSM \cite{bib3} & 69.2\% & 24.3\% & 36\% & 50.5\% \\
\hline
LSI \cite{bib2} & 70.5\% & 26\% & 38\% & 52.6\% \\
\hline
Fine-tuned BERT \cite{bib9} & 34.6\% & 12.6\% & 18.5\% & 25.7\% \\
\hline
Req2Vec \cite{bib10} & 59\% & 18.5\% & 28.1\% & 41\% \\
\hline
Ours & 67.9\% & 24.3\% & 35.8\% & 50\% \\
\hline
\end{tabular}%
}
\caption{\centering Results on WARC(FRS)}
\label{tab:WARC(FRS)}
\end{table}

\begin{figure}[H]
    \centering
    \includegraphics[width=1\linewidth]{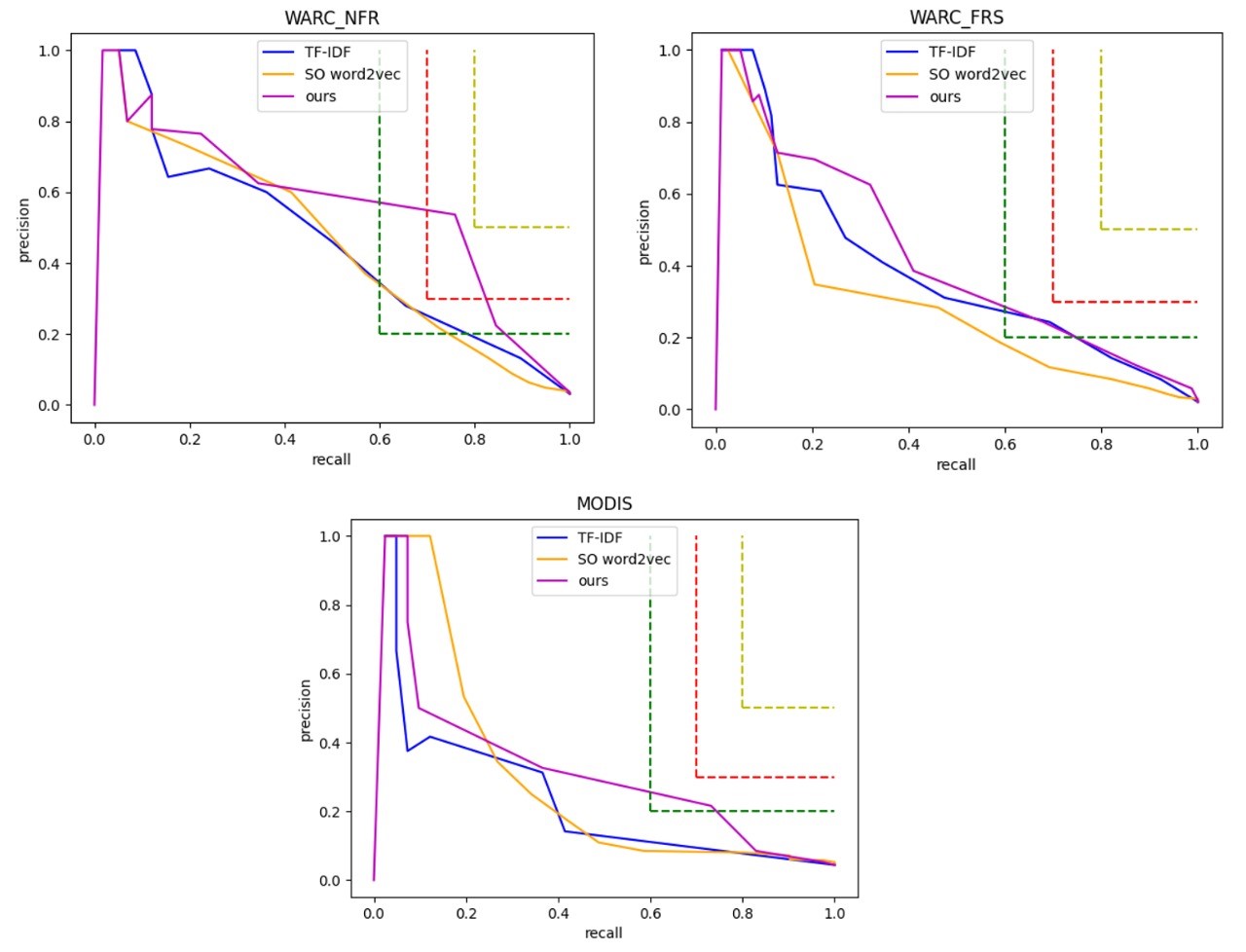}
    \caption{\centering precision/recall curves for the best methods on the three datasets.}
    \label{fig:results}
\end{figure}

Figure \ref{fig:results} demonstrates the recall/precision curves of the aforementioned methods on the datasets. To enhance clarity, we have selectively plotted the three best methods (which were VSM, Req2Vec, and Ours). The dotted lines in the colors green, red, and yellow, represent the Accepted, Good, and Excellent Regions, respectively.

\section{Discussion}
Our experimental results demonstrate that our approach surpasses existing state-of-the-art methods in the WARC(NFR) and MODIS datasets by 14.1\% and 18.4\% in terms of $F2$ while showing slightly lower performance in the WARC(FRS) dataset. Furthermore, our method has achieved better performance based on the criteria established by Hayes et al. \cite{bib18}. 

Specifically, our approach reached the Good level in the WARC(NFR) dataset (and was very close to entering the excellent level). Additionally, it attained an accepted level of results in the MODIS dataset. However, the performance level in the WARC(FRS) dataset remained unchanged when compared to the existing state-of-the-art methods.
 One key advantage of our approach over the traditional TF-IDF method is the incorporation of semantic information, which captures the relationships between words to better understand the relations between requirements. While Latent Semantic Indexing (LSI) also aims to achieve this by depending on the distributional theorem, it may be limited when working with small datasets like in the case of requirements linking problem, due to its reliance on pattern recognition. In contrast, our approach utilizes large amounts of data from embedding systems to capture and reflect this information more effectively. 
 
 On the other hand, although fine-tuned embedding models like FiBERT can represent the words as vectors very well, it wasn’t able to achieve good results due to the information loss incurred during the pooling stage, particularly mean pooling in the case of FiBERT. Lastly, Req2Vec which is a requirement-specific embedding, couldn’t improve the results, This can be attributed to the absence of a specific template or standardized structure within the requirements present in these datasets.

The effectiveness of our approach is influenced by various factors such as the domain of requirements and the diversity of expressions for a given term, leading to varying results across different datasets. For instance, comparing the WARC(NFR) and WARC(FRS) datasets, which focus on non-functional and functional requirements respectively, we observe significant differences in performance even though both datasets are from the same domain. This can be attributed to the narrower scope of non-functional requirements (e.g., security, availability, …) compared to functional ones, resulting in a greater variety of terms conveying similar meanings for non-functional requirements. This scenario presents an ideal environment for our approach to excel.

\section{Conclusion}
In this work we introduced a novel similarity function designed to complement Bag-of-Words (BOW) representation with Term Frequency-Inverse Document Frequency (TF-IDF) scoring. Our experimental evaluation shows promising results when dealing with the problem of cross-level requirements traceability. These findings provide evidence of the effectiveness of our proposed approach in mitigating the limitations associated with the utilization of cosine similarity in conjunction with BOW+TF-IDF.
In our future research, we aim to expand upon the concept of incorporating word embedding similarities by exploring the adjustment of sentence embeddings themselves. This extension holds potential for broader applications, including the utilization of these adjusted embeddings as features for training machine learning models and various other applications.

\section{Declarations}
\textbf{Conflict of Interest}: On behalf of all authors, the corresponding author states that there is no conflict of interest.
\newline
\textbf{Ethical Approval}: This article does not contain any studies with human participants performed by any of the authors.
\newline
\textbf{Funding}: This study has received no funding
\newline
\textbf{data availability}: all datasets used in this work are publicly available. for WARC(NFR) and WARC(FRS) at http://sarec.nd.edu/coest/datasets.html  and for MODIS http://promise.site.uottawa.ca/SERepository/datasets-page.html
\bibliography{sn-article}

\end{document}